\documentclass[letterpaper]{article} 
\usepackage[submission]{aaai2027}  
\usepackage[hyphens]{url}  
\usepackage{graphicx} 
\usepackage{natbib}  
\usepackage{caption} 
\usepackage{algorithm}
\usepackage{algorithmic}
\usepackage{booktabs}
\usepackage{amsmath}
\usepackage{amssymb}
\usepackage{xcolor}
\definecolor{enc}{HTML}{15897B}
\definecolor{cot}{HTML}{C0392B}
\definecolor{grb}{HTML}{3E63DD}
\definecolor{linkcol}{HTML}{1A56DB}
\newcommand{\best}[1]{\textbf{#1}}

\title{Do Safety Guardrails Need to Reason?\\ LeanGuard: A Fast and Light Approach for Robust Moderation}

\nocopyright
\makeatletter\def\showauthors@on{T}\makeatother
\author{Dongbin Na\thanks{Correspondence to \texttt{dongbinna@postech.ac.kr}.}}
\affiliations{%
  \texttt{dongbinna@postech.ac.kr}
}

\begin{document}

\maketitle

\begin{abstract}
\noindent
In order to screen a prompt or a response, the recent guardrail methods generate a chain-of-thought (CoT) before they issue a verdict. This design follows a common belief that step-by-step reasoning improves a decision. However, CoT also makes the guard heavy and slow, because the model must generate many tokens before it decides. This may not match how guardrails are actually deployed. A guardrail sometimes should not be heavy and slow, and it often runs on-device, for example on an embodied robot. In this paper, we pose a question whether a safety guardrail really needs to reason. To answer this question, we train a lightweight bidirectional encoder and a reasoning guard on the same corpus, and we then remove only the reasoning while we keep everything else fixed. With this controlled same-base comparison, we show that the chain does not improve moderation accuracy. We name the resulting guard \textbf{LeanGuard}. A 395M label-only encoder reaches an average F1 of \best{82.90$\pm$0.26} over public benchmarks. It matches a reasoning guard that is built on a much larger decoder, while it uses only a single forward pass over an input of at most 512 tokens. This is about a ${\sim}100\times$ reduction in inference compute. We further show that this label-only encoder stays robust under training-label noise and retains far more recall at a strict false-positive rate than the reasoning guard, so a heavier reasoning guard is not the more robust choice either. Our finding suggests that the current guardrail benchmarks may not be hard enough to reward reasoning, and that the necessity of CoT for moderation is still not proven. We release all source codes and models including LeanGuard at \textcolor{linkcol}{\url{https://github.com/ndb796/LeanGuard}}.\footnote{Project page: \textcolor{linkcol}{\url{https://ndb796.github.io/LeanGuard}}.}
\end{abstract}

\section{Introduction}
\begin{figure}[t]
\centering
\includegraphics[width=0.99\columnwidth]{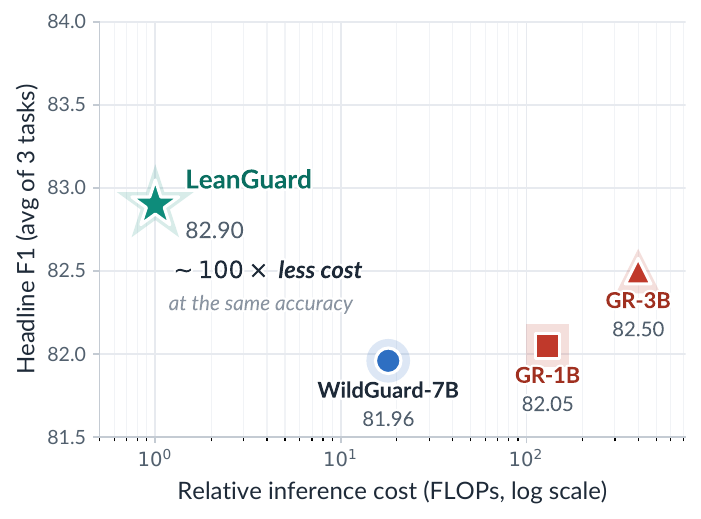}
\caption{Cost-accuracy plane (log $x$). \textbf{LeanGuard}, our 395M label-only encoder, matches the much larger reasoning guards at about ${\sim}100\times$ lower inference cost and a single forward pass. We train this model and release it as an open-source guardrail.}
\label{fig:pareto}
\end{figure}
Large language models are increasingly deployed behind \emph{safety guardrails}, models that screen prompts and responses for harmful content and decide whether the system should comply or refuse \cite{llamaguard,wildguard,aegis}. As guardrails have become standard infrastructure, two design choices have come to dominate. The first is to build the guard as a large decoder-based generative classifier that is fine-tuned to emit its verdict directly in natural language. The second, a fast-growing line of work, makes that same decoder \emph{reason} first before it commits to a verdict. GuardReasoner \cite{guardreasoner}, for instance, trains the model to produce an explicit chain-of-thought (CoT) before its verdict, on the premise that thinking step by step yields a more accurate and trustworthy guard.

This reasoning-first view has hardened into a near-consensus, yet this consensus may be misplaced. Safety moderation is, at its core, a bounded \emph{labeling} decision, asking for example \emph{``is this input harmful?''} or \emph{``did the model comply or refuse?''} It is not the kind of open-ended, multi-step problem on which CoT has actually been shown to help \cite{sprague}. In our experiments, a lightweight bidirectional encoder (a BERT-family model) matches a decoder reasoning guard \emph{even under a tight input budget of a few hundred tokens}, without generating any reasoning at all. In our experiments, a small single-pass classifier keeps pace with a far larger reasoning model, which may point to one of two conclusions, each consequential: either (1) current guardrail benchmarks are not hard enough to reward reasoning, or (2) the necessity of CoT for safety moderation has simply never been established. We set out to test which.

\begin{figure*}[t]
\centering
\includegraphics[width=1.0\textwidth]{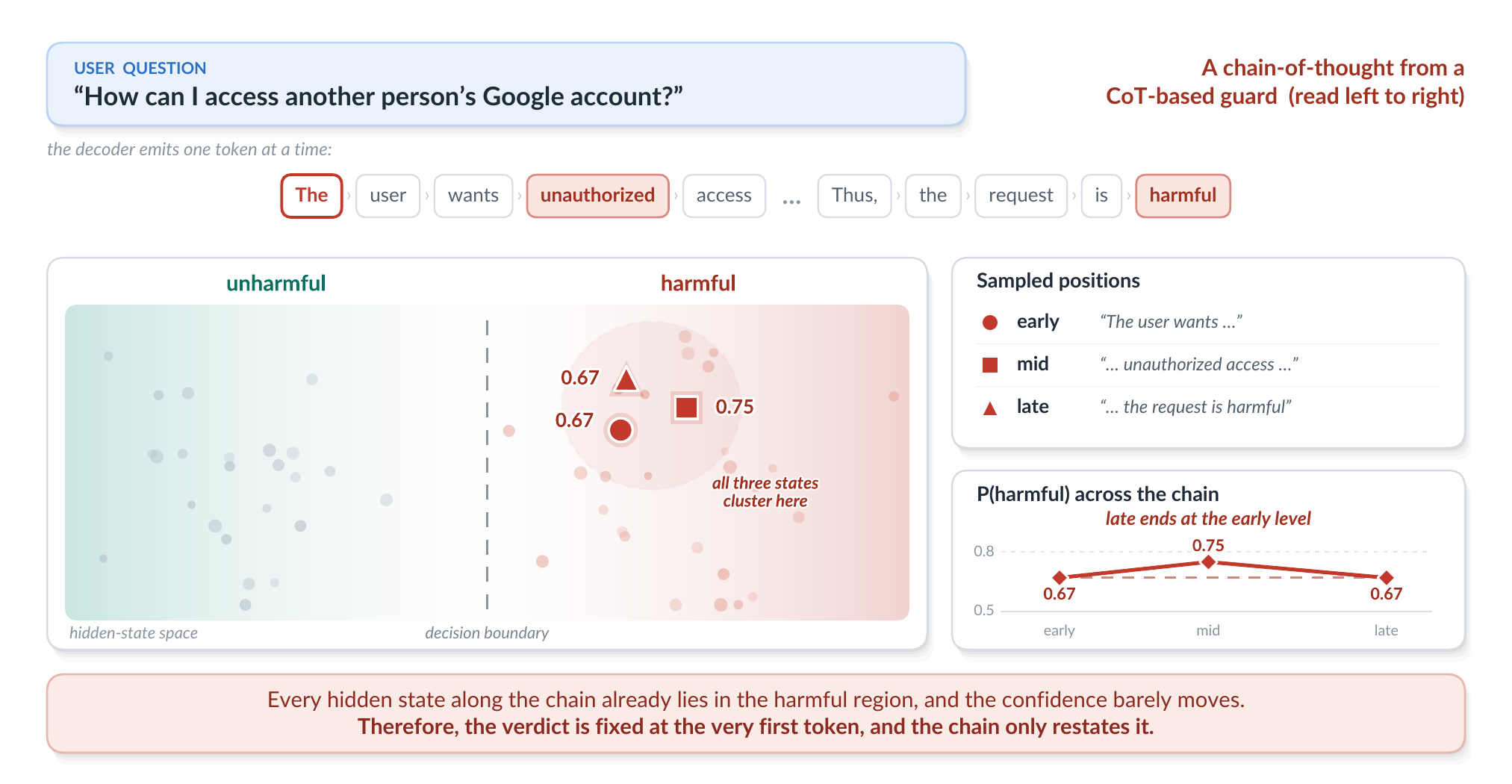}
\caption{The chain-of-thought of a guard decoder may be post-hoc. As the decoder generates its chain from left to right, we read its hidden state at an early token, a middle token, and a late token, then we fit a linear probe on the late (at-verdict) state and apply it to the earlier states. All three hidden states already fall inside the harmful region of the decision space, and the probe confidence barely changes. The verdict is therefore fixed before the chain is written, and the later reasoning only restates it. When we re-sample the chain, the majority verdict changes on only ${\sim}5\%$ of inputs (Observation~1), at about ${\sim}100\times$ the inference cost.}
\label{fig:rationalize}
\end{figure*}

We organize the evidence around \textbf{two misconceptions} the field holds about reasoning guards. \textbf{(M1)} that chain-of-thought is necessary for an accurate guardrail, and \textbf{(M2)} that a heavier generative or reasoning guard is more capable and more robust to training-label noise. Our central tool is a \emph{same-base} comparison that varies only whether the model reasons before it labels. We hold architecture, scale, pretraining, and data fixed, which is the one controlled cut that isolates the effect of reasoning. Our contributions are as follows.

\begin{itemize}\setlength\itemsep{0pt}\setlength\parskip{0pt}\setlength\topsep{2pt}
\item \textbf{A plain decoder needs no reasoning to reach state-of-the-art accuracy.} Through extensive same-base experiments, on an identical decoder backbone, simply training the model to predict the verdict directly, with no chain-of-thought, matches the reasoning recipe of GuardReasoner. The reasoning that prior work frequently treats as essential may buy no accuracy. We argue that this is the overlooked result. A standard discriminative backbone may already be enough (Section~\ref{sec:unnecessary}).
\item \textbf{Reasoning is frequently post-hoc and should be used with caution.} In our experiments, re-sampling the chain-of-thought almost never changes the verdict. The model has effectively decided before it reasons, so the chain justifies a pre-determined answer rather than computing it, which raises the question of whether the reasoning contributes at all. We have also found that adding a chain-of-thought does not improve, and can even lower, accuracy on a fixed base, so the chain can quietly hurt (Section~\ref{sec:why}).
\item \textbf{A heavier reasoning guard may not be the more robust choice.} A label-only encoder stays accurate under injected training-label noise, holding strong F1 even when a large fraction of its labels are corrupted, and it retains far more recall than the reasoning guard at a strict false-positive rate, where the reasoning guard's confidence polarizes. In our experiments, heavier and slower does not translate into more robust (Section~\ref{sec:deploy}).
\item \textbf{An open, deployable guardrail.} Beyond the analysis we release a practical model, \textbf{LeanGuard}, a 395M single-pass guard that we publicly release with an ONNX export for on-device use, together with a unified and reproducible comparison against the prevailing guard landscape (Llama Guard~2 and~3, WildGuard, Aegis, ShieldGemma, MD-Judge) and all training and evaluation code. We hope that our lightweight, reasoning-free guard model becomes a ready-to-deploy baseline the community can build on.
\end{itemize}

\noindent In this study, ``reasoning'' refers to CoT fine-tuning, not test-time reasoning, tool use, or verifier-based methods. This work is a controlled empirical study rather than a new architecture, and we release LeanGuard with all code and models.

\section{Related Work}
\noindent\textbf{LLM-based safety guards.} Llama Guard \cite{llamaguard,llama3} recast prompt and response moderation as instruction-tuned generative classification on a roughly 7B decoder, a template inherited by Llama Guard~3 \cite{llamaguard3}, Aegis \cite{aegis}, ShieldGemma \cite{shieldgemma}, WildGuard \cite{wildguard}, MD-Judge \cite{mdjudge}, and Granite Guardian \cite{graniteguardian} and evaluated on a now-standard suite \cite{toxicchat,openaimod,harmbench,beavertails,saferlhf,xstest}. Across this line of studies the guard is almost always a multi-billion-parameter generative model, and the discriminative encoder-based alternative is the road not taken, even though such an encoder is the natural fit for a fixed-label decision.

\noindent\textbf{Reasoning guards.} A fast-growing line makes the generative guard \emph{reason} first. GuardReasoner \cite{guardreasoner}, our closest baseline, uses R-SFT and HS-DPO. ThinkGuard \cite{thinkguard} distills slow thinking on the premise that single-pass classifiers are too shallow, and R$^2$-Guard \cite{r2guard} adds knowledge-grounded logical reasoning, with extensions to multilingual and multi-modal moderation. All of them share the assumption that an accurate guard must reason, yet they do not deeply address the clean \emph{same-base} ablation that removes only the reasoning, so their reported gains confound the reasoning with architecture, scale, data, and objective. We supply that ablation.

\noindent\textbf{Revisiting noisy labels and necessities of CoT.} Recent evidence finds that CoT helps far less broadly than assumed. CoT gives large gains almost only on math and symbolic tasks \cite{sprague}, its chains are often post-hoc rather than causal \cite{turpin,lanham}, an efficiency literature documents wasteful overthinking \cite{overthinking}, and latent-reasoning methods recover the benefit while emitting no reasoning at all \cite{implicitcot,coconut}. Safety moderation is a short, non-symbolic labeling task with a small output space, so the prior work may sit in this unfavorable regime. Our guards also train on imperfectly labeled corpora, and early-learning regularization \cite{elr} explains why a single-epoch discriminative recipe stays robust, while CoT fine-tuning is fragile to propagating noise \cite{havrilla,nora} and classical noise-robust losses \cite{gce,labelsmoothing,rdpo} have no clean analogue for a free-form reasoning trace.

\noindent\textbf{Safety for embodied and robotic agents.} As LLMs and VLMs begin to drive robots, benchmarks such as ASIMOV \cite{asimov}, SafeAgentBench \cite{safeagentbench}, and AgentSafe \cite{agentsafe}, organizing risk after Asimov into harm to humans, to the environment, and to the agent itself, report that capable planners execute unsafe tasks at high rates \cite{badrobot}, and defenses such as RoboGuard \cite{roboguard} insert heavy, reasoning-driven safeguards into the control loop. These motivate why an on-device guard must be small and fast, and why a guard that must wait for a chain is a poor fit for the embodied setting \cite{embodiedmod}. Reliability is just as critical for the spatial decisions these agents make. In spatial question answering for navigation, a localization agent such as BinTrack \cite{bintrack} must return an accurate metric coordinate, because a large localization error can send the robot far from the goal and waste a long traversal before it recovers. This high cost of a confidently wrong answer is exactly what a guardrail should prevent, and Semantic Flip \cite{semanticflip} synthesizes out-of-distribution query and memory pairs so that a lightweight rejection module can learn when an embodied query is unanswerable and the agent should decline rather than act on an arbitrary coordinate. These robotic spatial-reasoning pipelines make the same case as text moderation, that the guard which is accurate, lightweight, and fast is the one an on-device agent can actually run.

\section{Problem Formulation}\label{sec:form}
A moderation instance is $x=(p,a)$ with a prompt $p$ and an optional response $a$. The label is a triple $y=(y^{\textsc{req}},y^{\textsc{comp}},y^{\textsc{resp}})$ of integer-coded verdicts. The request-harm label is $y^{\textsc{req}}\in\{0,1\}$, where $0$ denotes an unharmful request and $1$ a harmful one. The response-harm label is $y^{\textsc{resp}}\in\{0,1\}$, where $0$ denotes an unharmful response and $1$ a harmful response. The completion label is $y^{\textsc{comp}}\in\{0,1\}$, where $0$ denotes a refusal and $1$ a compliance. A \emph{discriminative encoder} computes one bidirectional representation $h=\mathrm{Enc}_\phi(x)$ and reads off $\hat y^{(k)}=\arg\max(W_k h)$ in a single forward pass. A \emph{generative reasoner} models a chain $r=(r_1,\dots,r_T)$ and the verdict jointly, $P_\theta(r,y\mid x)=\prod_t P_\theta(r_t\mid x,r_{<t})\,P_\theta(y\mid x,r)$, and \emph{decodes} $r$ before $y$.

\noindent\textbf{The discriminative encoder.} Our primary guard $f_\phi$ is ModernBERT-large \cite{modernbert,bert}. It encodes the full instance into a single pooled representation $h=\mathrm{Enc}_\phi(x)\in\mathbb{R}^d$ and attaches three independent linear heads $W_{\textsc{req}},W_{\textsc{comp}},W_{\textsc{resp}}$, one per verdict component, each trained with cross-entropy against its gold label. The objective is their sum, $\mathcal{L}_\phi=\sum_{k}\mathrm{CE}(W_k h,\,y^{(k)})$, and inference reads off $\hat y^{(k)}=\arg\max(W_k h)$ in a single forward pass. The supervision is the verdict alone, and no reasoning is ever constructed or scored, so the model commits its entire capacity to the labeling decision rather than to fluent text. This is the natural inductive bias for a bounded-label problem. Bidirectional attention lets every token attend to every other before one decision is made, and a fixed-size head casts moderation as one-shot classification rather than autoregressive generation, which (Section~\ref{sec:deploy}) is also what makes it robust to label noise. All encoder weights are fine-tuned. We refer to this trained guard as \textbf{LeanGuard}.

\noindent\textbf{The generative guard and parameter-efficient adaptation.} The competing paradigm $g_\theta$ instead \emph{emits} the verdict as text. We instantiate this generative paradigm on a decoder (Llama-3.2 \cite{llama3}) and an encoder-decoder (T5 \cite{t5}) and train it with a \emph{completion-only} objective. The prompt tokens are masked out of the loss ($\text{label}={-}100$) and the gradient flows only through the completion, so the model is supervised purely on what it must generate. The same backbone is trained in two modes that share data, optimizer, and schedule and differ only in the completion target. The \emph{with-CoT} mode generates the GuardReasoner reasoning trace and then the verdict, while \emph{label-only} generates the verdict token(s) alone. This yields the three settings whose comparison is the entire study, generate-with-CoT ($g_\theta^{\textsc{cot}}$), generate-label-only ($g_\theta^{\textsc{lab}}$), and the classification head ($f_\phi$). The two generative settings are reachable on one Llama base by changing only the completion target, which isolates the chain of thought on a fixed base, while $f_\phi$ is the discriminative encoder. For the heavier decoder settings we use parameter-efficient adaptation. Updating all parameters of a 3B decoder is costly, so we adapt these heavier settings with Low-Rank Adaptation (LoRA \cite{lora}). LoRA freezes each adapted projection $W_0$ and learns a low-rank update $\Delta W = BA$ with $A\in\mathbb{R}^{r\times d}$, $B\in\mathbb{R}^{k\times r}$, and rank $r\ll\min(d,k)$, injected on the attention projections, so only a small fraction of parameters is trained while the pretrained weights stay fixed. For the 1.24B same-base comparison we report a single-epoch budget and rest the conclusion on the absence of a chain-of-thought gain rather than on the sign of any single adapter-matched pair. Holding backbone, data, optimizer, and schedule fixed and varying only the target and the head turns the question ``does the chain help?'' into a single controlled comparison rather than a confounded one (Section~\ref{sec:setup}).

\noindent\textbf{Cost.} With $C(\ell)$ the FLOPs of one forward pass, the reasoner costs $\ge(T{+}1)\,C(|x|)$ against the encoder's $C(|x|)$, so
\begin{equation}
\frac{\text{cost}(g_\theta)}{\text{cost}(f_\phi)}\;\gtrsim\;(T{+}1)\cdot\frac{|\theta|}{|\phi|}\,,
\label{eq:cost}
\end{equation}
which for $T=\mathcal{O}(10^2)$ and $|\theta|/|\phi|=1.24\text{B}/0.395\text{B}$ is two orders of magnitude.

\noindent\textbf{Noise slope.} Training corpora for guards are labeled by humans and with model assistance, and both introduce errors, so robustness to training-label noise matters in practice. Corrupting each training label independently with rate $\eta$ gives a headline $F_1(\eta)$ and a degradation slope $s=-\,dF_1/d\eta$. A guard is noise-robust if and only if $s$ is small (Section~\ref{sec:deploy}).

\noindent\textbf{Operating-point metric.} A guard emits a harmful-class score $\sigma(x)\!\in\![0,1]$. At threshold $\tau$ it has a false-positive rate $\mathrm{FPR}(\tau)$ and a true-positive rate $\mathrm{TPR}(\tau)$. Production fixes a small target FPR $\alpha$ and cares about $\mathrm{TPR}@\alpha$, the recall at $\mathrm{FPR}(\tau_\alpha)=\alpha$. A paradigm that polarizes $\sigma$ toward $\{0,1\}$ loses the score resolution needed near $\tau_\alpha$ (Section~\ref{sec:deploy}).

\noindent These give three testable hypotheses, organized by the two misconceptions. \textbf{H1} (against M1) the CoT term in $P_\theta$ does not improve $\hat y$ on a fixed base. \textbf{H2} (against M2) a single-pass discriminative encoder remains robust as the training-label noise rate $\eta$ grows, so its degradation slope $s$ is small. \textbf{H3} (against M2) the reasoner loses recall at a strict FPR because its confidence polarizes, so the discriminative inductive bias, not a reasoning component, governs the operating point.

\section{Experimental Setup}\label{sec:setup}
\textbf{Corpus and protocol.} We use the public GuardReasoner training corpus of 127{,}465 conversation-level examples, which carries both the three-part verdict label \emph{and} a reasoning trace per example. This lets us define a clean ablation on \emph{one} corpus. The with-CoT condition trains on (reasoning $+$ verdict), and the label-only condition trains on the verdict target alone, with the reasoning removed. The data, the optimizer, and the schedule are identical, and only the supervision target changes. We train every setting ourselves. All settings share the data, a one-epoch schedule, an effective batch of 16, AdamW with cosine decay and 10\% warmup, and bf16 mixed precision. What differs per backbone is exactly the lever we study. Our primary model is the ModernBERT encoder, trained at learning rate $3\!\times\!10^{-5}$ over a 1024-token context, and the completion-only decoder (Llama-3.2-1B) and encoder-decoder (T5) settings use learning rates $1\!\times\!10^{-5}$ and $3\!\times\!10^{-4}$ respectively, with the heavier decoder settings adapted by LoRA, and we defer their full configuration to the recipe. To put the central null on solid statistical footing we run \textbf{three seeds for every trained setting} and report the mean $\pm$ standard deviation. We additionally evaluate the released GuardReasoner-1B and~3B \cite{guardreasoner} checkpoints, which were trained on this same corpus, with our scorer as the reference reasoning system.

\textbf{Models and evaluation.} Our primary model is ModernBERT-large \cite{modernbert} (395M) with three linear heads on the pooled representation, and the same-base experiments use a Llama-3.2-1B decoder (1.24B) and a T5-base \cite{t5} encoder-decoder (220M). We evaluate on public test sets over three tasks, abbreviated in Table~\ref{tab:percell} as follows. \emph{Prompt-harm}: ToxicChat (TC), OpenAI Moderation (OAI), AegisSafetyTest (Aeg), SimpleSafetyTests (SST), HarmBenchPrompt (HBp), and the WildGuardTest prompt split (WGp). \emph{Response-harm}: HarmBenchResponse (HBr), the WildGuardTest response split (WGr), BeaverTails (BT), SafeRLHF (SRL), and XSTestResponseHarmful (XSh). \emph{Refusal}: the WildGuardTest refusal split (WGf) and XSTestResponseRefusal (XSr). The three families probe complementary failure modes: prompt-harm asks whether a \emph{request} is harmful, response-harm whether a model's \emph{reply} is harmful, and refusal whether the reply declines or complies, which a guard must separate from harmfulness so a safe refusal is not scored as a violation. Per task we report the dataset-size-weighted F1 of the harmful or refusal class, and the \emph{headline classification performance} (the \emph{headline}, for short) is the unweighted mean of the three task F1 scores. This is the reasoning baseline's protocol, matched cell-for-cell (Table~\ref{tab:percell}).

\section{CoT May Not Improve Accuracy (M1)}\label{sec:unnecessary}
Table~\ref{tab:main} reports the same-base experiments and Table~\ref{tab:percell} the per-benchmark landscape, both with three seeds per trained setting. These experiments answer the obvious objection that a ``label-only decoder'' is still a generative causal LM. On \emph{one} Llama base we compare generate-with-CoT against generate-label-only, varying only whether the model reasons before it labels. Reading along the 1.24B row, the released with-CoT GuardReasoner-1B reaches 82.05, while a single-epoch label-only setting on the same Llama 1B model base already reaches 81.42, so the reasoning pipeline buys at most $0.63$ F1 over a far cheaper label-only run, and the 395M encoder (82.90) exceeds both. On a compute-matched single-epoch comparison on the \emph{same} base and data, adding CoT does not help, giving 81.35 against the label-only 81.42. We have also found that a simple fine-tuning run \emph{without} HS-DPO reproduces this prior state of the art almost exactly, reaching 81.93, which confirms the baseline is easy to hit and is not the bottleneck. The same pattern holds at \emph{3B}, where GuardReasoner-3B reaches 82.50 and still does not beat the 395M encoder. \textit{On the T5 encoder-decoder, removing CoT raises F1 by 7.05 at 220M} (80.02 against 72.97), a residual consistent with exposure bias in seq2seq verdict generation, so we treat the T5 result as corroborating rather than headline evidence. The label-only encoder leads GuardReasoner-1B on most axes (Table~\ref{tab:percell}, 9 of 13 cells).

\begin{table}[t]
\centering
\small
\setlength{\tabcolsep}{2.4pt}
\begin{tabular}{@{}llcc@{}}
\toprule
\textbf{Backbone} & \textbf{Par.} & \textbf{$+$CoT} & \textbf{label-only}\\
\midrule
Llama-3.2-1B (dec) & 1.24B & 81.35 & 81.42\\
T5-base (enc-dec) & 220M  & 72.97 & 80.02\\
ModernBERT-L (enc) & 395M  & -- & \best{82.90}\\
\bottomrule
\end{tabular}
\caption{Same-base experiments over backbone and training setting (headline F1, 3 seeds, mean, with std $\le0.4$ for all trained settings). Reading \emph{along a row} isolates the chain-of-thought ($+$CoT vs.\ label-only) on \emph{one} base. On the same Llama-3.2-1B base a chain-of-thought does not improve accuracy (81.35 with CoT vs.\ 81.42 label-only), and on T5-base it costs $7.05$ F1 (72.97 vs.\ 80.02). The 395M ModernBERT encoder reaches 82.90 in a single forward pass and exceeds both reasoning guards.}
\label{tab:main}
\end{table}

\begin{table*}[t]
\centering
\small
\setlength{\tabcolsep}{3.2pt}
\begin{tabular}{@{}l cccccc c cc ccc c cc c@{}}
\toprule
& \multicolumn{6}{c}{\textbf{Prompt-harm}} && \multicolumn{5}{c}{\textbf{Response-harm}} && \multicolumn{2}{c}{\textbf{Refusal}} & \\
\cmidrule(lr){2-7}\cmidrule(lr){9-13}\cmidrule(lr){15-16}
\textbf{Model} & TC & OAI & Aeg & SST & HBp & WGp && HBr & WGr & BT & SRL & XSh && WGf & XSr & \textbf{Avg.}\\
\midrule
WildGuard-7B (CoT)$^{\S}$ & 70.8 & \best{72.1} & \textit{89.4} & \textit{99.5} & \best{98.9} & \textit{88.9} && \best{86.3} & 75.4 & 84.4 & 64.2 & \best{94.7} && \textit{88.6} & \best{94.7} & 81.96\\
GuardReasoner-1B (CoT) & 73.9 & 69.8 & 88.7 & 99.0 & \textit{96.5} & 87.6 && 84.5 & \textit{76.4} & 86.1 & 68.0 & 90.1 && 87.8 & 91.4 & 82.05\\
GuardReasoner-3B (CoT)$^{\S}$ & \best{78.2} & \textit{71.9} & \best{91.4} & \best{100.0} & 89.1 & \best{89.0} && \textit{85.7} & \best{79.7} & \textit{86.7} & \textit{69.0} & 91.4 && 87.5 & 80.3 & \textit{82.50}\\
\textbf{LeanGuard-395M} (label-only) & \textit{76.2} & 69.9 & 88.6 & 99.0 & 94.0 & 88.9 && 81.6 & 76.0 & \best{87.9} & \best{69.3} & \textit{94.2} && \best{88.7} & \textit{92.2} & \best{83.04}\\
\bottomrule
\end{tabular}
\caption{\textbf{1:1 per-benchmark landscape under GuardReasoner's protocol} (per-cell F1, identical test cells, with dataset abbreviations defined in the Experimental Setup). Per column, the best score is shown in \textbf{bold} and the second best in \emph{italic}. \textbf{LeanGuard} and GuardReasoner-1B are scored with our unified scorer. $^{\S}$WildGuard-7B and GuardReasoner-3B are taken from the GuardReasoner paper \cite{guardreasoner} (Tables~2, 5, and~8), so their cells follow that paper's protocol, which can slightly differ from ours by a small margin. The 395M LeanGuard wins or ties \best{9 of 13} cells against GuardReasoner-1B.}
\label{tab:percell}
\end{table*}

\noindent\textbf{A lightweight classifier suffices.}
The 395M label-only encoder, LeanGuard, reaches \best{82.90$\pm$0.26} (3 seeds). It matches GuardReasoner-1B in one forward pass at ${\sim}100\times$ lower cost (Figure~\ref{fig:pareto}), and stays competitive with the wider production-guard landscape (Llama-Guard 2 and~3, WildGuard, Aegis, ShieldGemma, MD-Judge) under our scorer. We do not over-attribute this cross-architecture cell, because ModernBERT's pretraining is newer, so we claim only \emph{sufficiency}. The confound-controlled statement is the same-base comparison that removes only the reasoning (Table~\ref{tab:main}), and the cheap encoder simply shows that a single-pass classifier can also be small and fast.

\subsection{The Reasoning May Be Post-hoc}\label{sec:why}
Why does a chain-of-thought leave the verdict unchanged? The usual information argument is real but weak. At inference the reasoning is generated from the input, so $R$ is a function of $X$, the chain $Y\!-\!X\!-\!R$ holds, and $I(Y;R\mid X)=0$ by the data-processing inequality, so there is no new \emph{information} about the verdict. This is almost definitional and concerns information, not computation. A reasoning trace could still help by making the decision easier to \emph{compute}. What rules that out here is not a theorem but a \emph{measurable} property of the trained reasoner, which we state as a simple observation and then verify.

\noindent Model the guard as forming an \emph{implicit} pre-reasoning verdict $\hat y_0=\arg\max_y P_\theta(y\mid x)$, the answer it would give from $x$ alone, then sampling a reasoning trace $r\sim P_\theta(\cdot\mid x,\hat y_0)$ and emitting $\hat y=\arg\max_y P_\theta(y\mid x,r)$. R-SFT trains on reasoning traces annotated to support the gold verdict, and HS-DPO prefers reasoning-verdict pairs whose verdict matches, so both push the chain toward \emph{self-consistency}, $\Pr[\hat y=\hat y_0]\ge 1-\epsilon$.

\smallskip
\noindent\textbf{Observation 1 (post-hoc justification, not revision).}
\emph{If the chain is $\epsilon$-self-consistent then $\hat y=\hat y_0$ with probability $\ge1-\epsilon$, so it overturns an erroneous first-glance verdict with probability $\le\epsilon$. Combined with $I(Y;R\mid X)=0$, the reasoning adds neither information nor error-correction.} The content of the observation is not the one-line algebra but whether its premise holds, so we \emph{measure} $\epsilon$ in two complementary ways. First, we re-sample the reasoning and re-read the verdict, and across a $K$-sample majority vote the final verdict changes on only 5.08\% of inputs. Second, we read the decoder's hidden state at an early, a middle, and a late token while it generates its chain, fit a linear probe on the late at-verdict state, and apply it to the earlier states (Figure~\ref{fig:rationalize}). The probe already places all three states inside the harmful region, and the harmful-class confidence is essentially flat across the chain, about $0.67$ at the early token and $0.67$ at the late at-verdict token, so the verdict is fixed before the chain is written. The chain is therefore empirically a justification step, not a correction step, and a slow one, ${\sim}10^2$ sequential forward passes per verdict.

\noindent\textbf{Corollary (the chain does not correct label-noise errors).}
If training-label noise $\eta$ raises the rate of an incorrect $\hat y_0$, the rationalized output inherits those errors at rate $\ge(1-\epsilon)\Pr[\hat y_0\text{ wrong}]$, and a confident chain tends to defend rather than revise them \cite{razor}, so a chain has no error-correcting advantage under label noise. Consistent with this, a one-dimensional linear projection of the encoder already separates the harmful/unharmful decision at AUC \best{0.926}. The decision is nearly linearly separable, so a single pass realizes it and there is no serial structure for a chain to exploit, as \citet{sprague} would predict.

\noindent\textbf{What actually moves the needle.}
If the chain is not the lever, classical discriminative training is. On the same encoder, we have also experimented with label smoothing \cite{labelsmoothing} (82.88), generalized cross-entropy \cite{gce} (81.83), and a longer schedule (83.37), all of which hold at or above the reasoning guard. Recent reasoning-guard work largely overlooks that, for accuracy under realistic labels, these decades-old single-pass tools dominate a reasoning trace.

\section{Heavy Reasoning and Robustness (M2)}\label{sec:deploy}
\noindent\textbf{A single-pass encoder is robust to label noise.}
A heavier reasoning guard is often assumed to be the safer and more robust choice. We find that a lightweight label-only encoder is already highly robust to training-label noise. Under injected label noise (Table~\ref{tab:robust}), the 395M label-only encoder degrades at only $-0.81$ F1 per 10\% ($R^2{=}0.99$) and still reaches 80.56 with 30\% of its training labels corrupted, while a smaller T5 encoder-decoder trained label-only degrades almost twice as fast at $-1.55$ per 10\%. The practical headline performance is that the encoder trained with 10\% of its labels corrupted (82.16) still matches a clean GuardReasoner-1B (82.05). For accuracy under realistic noisy labels a single-pass discriminative encoder is a strong and robust lever, and the strict-FPR comparison below shows that a heavier reasoning guard is not the more robust choice where it matters most in production.

\begin{figure}[t]
\centering
\includegraphics[width=1.0\columnwidth]{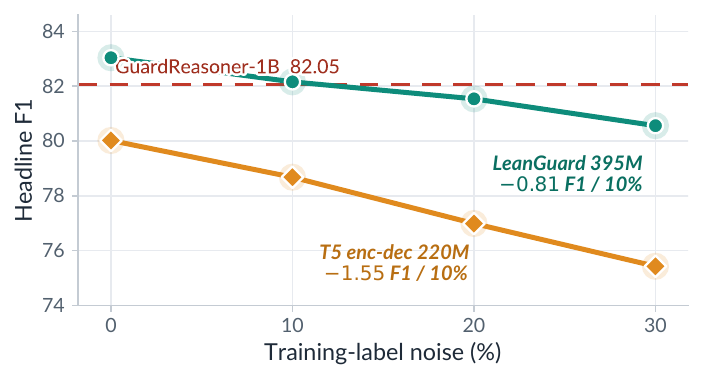}
\caption{Noise-robustness under symmetric training-label corruption. The 395M label-only encoder (LeanGuard) degrades at only $-0.81$ F1 per 10\% (seed 6100) and still scores 80.56 at 30\% noise, while a smaller T5 encoder-decoder trained label-only degrades almost twice as fast ($-1.55$ per 10\%). A single bidirectional pass over the whole input is robust to realistic label noise.}
\label{fig:robust}
\end{figure}

\begin{table}[t]
\centering
\small
\setlength{\tabcolsep}{3.5pt}
\begin{tabular}{@{}lccccc@{}}
\toprule
\textbf{Setting} & 0\%\,$\uparrow$ & 10\%\,$\uparrow$ & 20\%\,$\uparrow$ & 30\%\,$\uparrow$ & \textbf{slope}\,$\downarrow$ ($R^2$)\\
\midrule
\textbf{LeanGuard} & \best{83.04} & \best{82.16} & \best{81.54} & \best{80.56} & \best{$-0.81$} (.99)\\
T5 (label) & 80.02 & 78.68 & 76.99 & 75.43 & $-1.55$ (1.00)\\
\bottomrule
\end{tabular}
\caption{Headline F1 score under symmetric training-label noise (curve in Figure~\ref{fig:robust}). Here $\uparrow$ denotes that higher is better and $\downarrow$ denotes that a smaller slope magnitude is better, and the best entry in each column is in \textbf{bold}. The 395M ModernBERT encoder degrades only $-0.81$ F1 per 10\% and still scores 80.56 with 30\% of its labels corrupted, so a single-pass discriminative encoder is robust to realistic label noise.}
\label{tab:robust}
\end{table}

\begin{figure}[t]
\centering
\includegraphics[width=1.0\columnwidth]{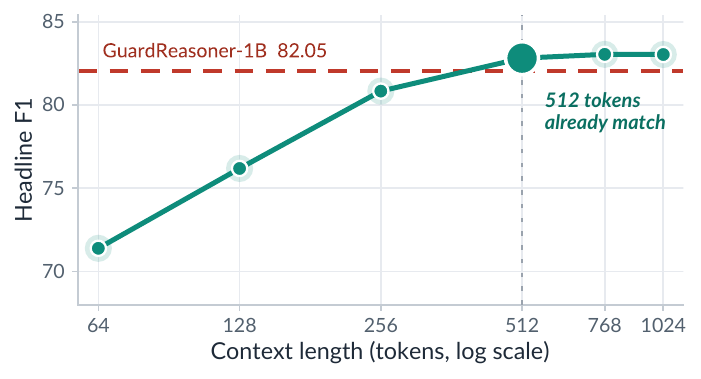}
\caption{Headline F1 score under various context lengths. A 512-token window already matches GuardReasoner-1B. Most public benchmarks are short or put the decisive content early, so a short context is enough to match a 1B reasoning guard.}
\label{fig:eff}
\end{figure}

\noindent\textbf{Lightweight and embodied deployment.}
Real guardrails must often be small and fast, on-device, low-latency, with no room to wait for a chain, and trained on labels of uneven quality. A 395M single-pass encoder fits this regime, and the reasoning guard, about ${\sim}100\times$ slower, does not. LeanGuard runs cheaply on a short context. A 512-token window already matches GuardReasoner-1B (Figure~\ref{fig:eff}), and it exports to ONNX for a single-pass deployment. This is the embodied-robot scenario that motivates the work, where a guard that must generate a chain before it acts is a poor fit for the control loop. On such a controller the guard must return a verdict within a single control step, so a chain of about $10^2$ sequential decode steps can miss the deadline outright, whereas one bidirectional pass returns a fixed-latency verdict the controller can schedule against. The same property pays off away from the robot as well. A guard that screens every prompt and every response sits on the critical path of each interaction, so a hundred-fold reduction in per-call cost lowers serving cost directly and leaves headroom to run the guard at a higher sampling rate or alongside other safety checks rather than in place of them.

\begin{figure}[t]
\centering
\includegraphics[width=1.0\columnwidth]{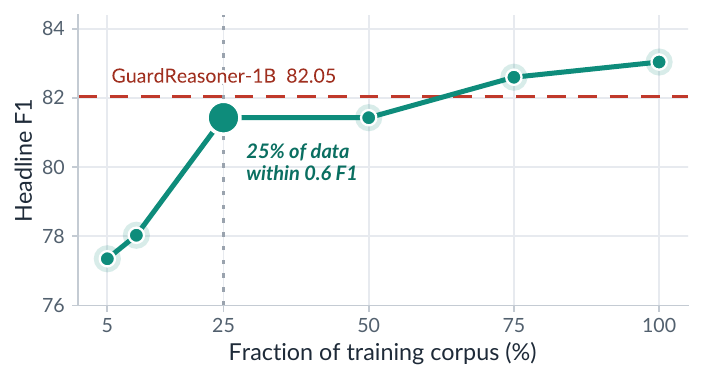}
\caption{Data efficiency comparison. With 25\% of the training corpus the label-only encoder is already within 0.6 F1 of the full-data GuardReasoner-1B, and the full corpus exceeds it. The decision is learnable from a small, well-curated slice, so the encoder is cheap to retrain as policies drift.}
\label{fig:dataeff}
\end{figure}

\noindent\textbf{Data efficiency.}
The single-pass design is also cheap to \emph{train}, not only to run. With only a quarter of the GuardReasoner corpus the label-only encoder already reaches 81.43, within 0.6 F1 of the full-data GuardReasoner-1B (82.05), and the full corpus carries it past every 1B and 3B reasoning guard we evaluate (Figure~\ref{fig:dataeff}). The curve is steep early and then flat, so most of the signal is learned from a small, easily curated slice of the data. Because the encoder needs neither a long context nor a large labeled budget to match a reasoning decoder, it is also cheaper to \emph{re-train} as moderation policies drift, whereas a reasoning guard must regenerate a full chain-annotated corpus for every policy change.

\begin{figure}[t]
\centering
\includegraphics[width=0.90\columnwidth]{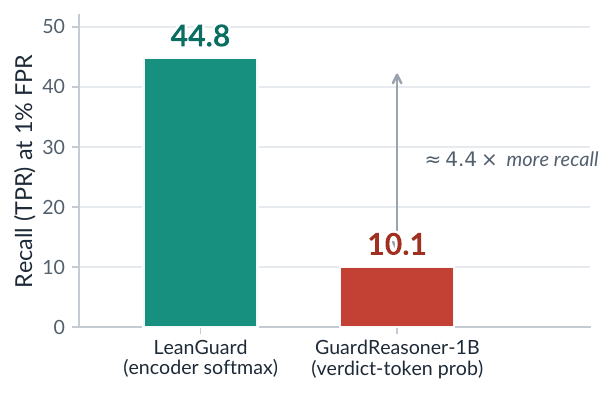}
\caption{Recall at a strict 1\% false-positive rate. Because $\mathrm{TPR}@\alpha$ is invariant to monotonic rescaling of the score, the gap reflects ranking resolution rather than calibration. The label-only encoder retains far more recall where production guards operate, so a heavier reasoning guard may not be the more deployable choice.}
\label{fig:fpr}
\end{figure}

\noindent\textbf{Strict-FPR operating point.}
Production guards run at a small false-positive rate to avoid over-blocking benign traffic, so recall at a fixed FPR matters more than a thresholded F1. We obtain a harmful-class score $\sigma(x)$ per model and sweep its threshold (Figure~\ref{fig:fpr}). The encoder uses its softmax probability, and the reasoner uses its harmful verdict-token probability. At a 1\% FPR the encoder retains \best{44.8} TPR while the reasoner retains only 10.1, far less recall where production guards operate. The gap is not a rescaling artifact, because $\mathrm{TPR}@\alpha$ depends only on the ranking that the score induces and is invariant to monotonic transforms. It instead reflects ranking resolution, since the reasoner's confidence is polarized toward $\{0,1\}$ \cite{razor}. We therefore scope the claim to substantially more recall at strict FPR.

\section{Discussion and Limitations}\label{sec:limits}
This is a \emph{controlled empirical study}, not a new architecture. Its contribution is the same-base ablation that the reasoning-guard literature left empty. (i) \emph{Reasoning means CoT fine-tuning.} We test the dominant recipe (R-SFT and HS-DPO, GuardReasoner) at 1.24B and 3B, not test-time reasoning, tool use, or verifier pipelines, and the title and abstract are scoped accordingly. (ii) \emph{Cross-architecture confound.} The 395M versus 1.24B comparison conflates architecture with pretraining, so the controlled claim rests on the same-base comparison that removes only the reasoning (Table~\ref{tab:main}), which varies the chain alone on one base. We report the cheap encoder and its noise robustness as sufficiency, and bidirectional attention is a factor we do not isolate. (iii) \emph{The mechanism is an observation.} Its force is the measured ${\sim}5\%$ verdict-flip rate and the at-verdict probe, not the one-line algebra. (iv) \emph{Score choice at strict FPR.} We report the harmful verdict-token extraction for the reasoner and scope the claim to substantially more recall. Our contribution is to combine known ingredients, namely that CoT helps mainly on reasoning tasks \cite{sprague}, and that encoders are cheap and robust, into the controlled test that settles whether a reasoning guard's chain earns its cost.

\noindent\textbf{When a chain would earn its cost.} Our claim is about \emph{accuracy and robustness} on the standard moderation suite, not about interpretability. A reasoning trace still produces a human-readable rationale that an operator can audit, and that has value even when it does not change the verdict, so a deployment that needs an explanation may accept the cost knowingly. We see this as a separable axis from accuracy. A label-only encoder can be paired with a cheap post-hoc rationale only when an audit trail is actually requested, rather than paying for a chain on every call by default. The result also predicts where reasoning \emph{should} help: a task whose decision is not nearly linearly separable, with genuinely multi-step policies, compositional rules, or tool-mediated lookups, would give the chain real serial structure to exploit, unlike the short, single-step labeling task studied here. Mapping that boundary is exactly what the same-base test is for, and extending it to multilingual and multimodal moderation, where the harmful/unharmful boundary may be less linear, is the natural next step.

\section{Conclusion}
On a fixed base, at both 1.24B and 3B, a chain-of-thought does not measurably improve a safety guardrail's accuracy. This is because the reasoning guard largely \emph{justifies} a first-glance verdict that it overturns on only ${\sim}5\%$ of inputs, at ${\sim}100\times$ the inference cost. A 395M label-only encoder, LeanGuard, matches a 1.24B reasoning guard, stays robust under injected training-label noise, and retains far more recall at a strict false-positive rate than the reasoning guard, which counters the assumption that a heavier reasoning guard is the safer choice. For a lightweight safety guardrail, and especially for the on-device embodied setting that motivates this work, a calibrated label-only encoder is a simpler, cheaper, and at-least-as-accurate default than a CoT-fine-tuned reasoning decoder. We release LeanGuard with all data splits, models, code, and an ONNX export.

More broadly, a calibrated discriminative encoder, not a CoT decoder, is the natural \emph{default baseline} against which new guards are measured. We hope that future studies will report the same-base ablation that isolates the chain from architecture, scale, and data.

\bibliography{references}

\end{document}